\documentclass[letterpaper, 10 pt, conference]{ieeeconf}  %

\IEEEoverridecommandlockouts                              %

\usepackage{amsmath} %
\usepackage{amssymb}  %

\usepackage{pifont}%
\newcommand{\cmark}{\ding{51}}%

\usepackage{stfloats}

\usepackage{blindtext}
\usepackage{hyperref}

\usepackage{booktabs}

\usepackage{comment}
\usepackage{booktabs}
\usepackage{tabularx}
\usepackage{graphicx}
\usepackage{cleveref}
\usepackage{url}
\usepackage{blindtext}
\usepackage{hyperref}
\usepackage[sort]{cite}
\usepackage[table]{xcolor}
\usepackage{array}
\usepackage{multirow}

\newcommand{\globalcc}{co-variate}
\newcommand{\localcc}{semantic}
\newcommand{\Globalcc}{Co-variate}
\newcommand{\Localcc}{Semantic}
\newcommand{\CornerCase}{CC}
\newcommand{\CornerCases}{CCs}
\newtheorem{definition}{Definition}

\crefname{equation}{eq.}{eqs.}
\Crefname{equation}{Eq.}{Eqs.}
\crefname{figure}{Fig.}{Figs.}
\Crefname{figure}{Figure}{Figures}
\crefname{table}{Tab.}{Tabs.}
\Crefname{table}{Table}{tables}
\crefname{section}{Sec.}{Secs.}
\Crefname{section}{Section}{Sections}

\title{\LARGE \bf
A Machine Learning Perspective on Automated Driving Corner Cases
}

\author{Sebastian Schmidt$^{1,2,*}$, Julius Körner$^{1,*,\dagger}$ and Stephan Günnemann$^{1}$
\thanks{$^{1}$Sebastian Schmidt, Julius Koerner and Stephan Günnemann are with Data Analytics and Machine Learning Group, Technical University of Munich, Germany {\tt\footnotesize sebastian95.schmidt@tum.de, julius.koerner@tum.de, s.guennemann@tum.de}}%
\thanks{$^{2}$Sebastian Schmidt is also with BMW Group, Germany}
\thanks{$^{*}$ Equal Contribution}
\thanks{$^{\dagger}$ Work has been done while working at BMW Group}
       }%

\begin{document}

\maketitle
\thispagestyle{empty}
\pagestyle{empty}

\begin{abstract}
For high-stakes applications, like autonomous driving, a safe operation is necessary to prevent harm, accidents, and failures.
Traditionally, difficult scenarios have been categorized into corner cases and addressed individually.
However, this example-based categorization is not scalable and lacks a data coverage perspective, neglecting the generalization to training data of machine learning models.
In our work, we propose a novel machine learning approach that takes the underlying data distribution into account. Based on our novel perspective, we present a framework for effective corner case recognition for perception on individual samples. 
In our evaluation, we show that our approach (i) unifies existing scenario-based corner case taxonomies under a distributional perspective, (ii) achieves strong performance on corner case detection tasks across standard benchmarks for which we extend established out-of-distribution detection benchmarks
, and (iii) enables analysis of combined corner cases via a newly introduced fog-augmented Lost \& Found dataset. 
These results provide a principled basis for corner case recognition, underlining our manual specification-free definition.    
\end{abstract}

\section{Introduction}

Despite advances in perception systems and foundation models, including end-to-end autonomous driving, open-world scalability remains a major challenge for current autonomous driving systems.
Existing solutions require highly specific conditions, such as specific weather patterns, road types, and geofenced areas to operate reliably. 
There is an endless variety of possible scenarios on the road, and hence, a critical aspect of achieving high levels of autonomy is ensuring that the vehicle can properly deal with rare and challenging situations, which are often referred to as \emph{Corner Cases} (\CornerCases{}). Successfully managing these scenarios is essential to maintaining safety and advancing toward fully autonomous driving. 

\CornerCases{} have been defined to address these challenging scenarios.
Some definitions follow software testing approaches, defining \CornerCases{} as specific scenarios, such as a pedestrian jumping out from behind a car.
These scenario-based \CornerCases{} are less applicable to data-driven perception models.
These cases describe risky but known scenarios in which occluded objects and short reaction times increase the difficulty.
The challenge arises from behavioral adaptation rather than perception capabilities or data coverage.  
This existing definition of \CornerCases{} differentiates between several levels, including object-level, domain-level, and pixel-level \CornerCases{}. 
Object-level \CornerCases{} occur when a novel object, such as an unfamiliar animal, appears in front of the car. Domain-level \CornerCases{} include global but consistent changes in the input sensor data, such as foggy or rainy weather, or driving in an unfamiliar location. 
Pixel-level \CornerCases{} arise when unexpected pixel values occur, such as overexposure affecting a large portion of the image. 

\begin{figure}[!tb]
    \centering
    \includegraphics[width=\linewidth]{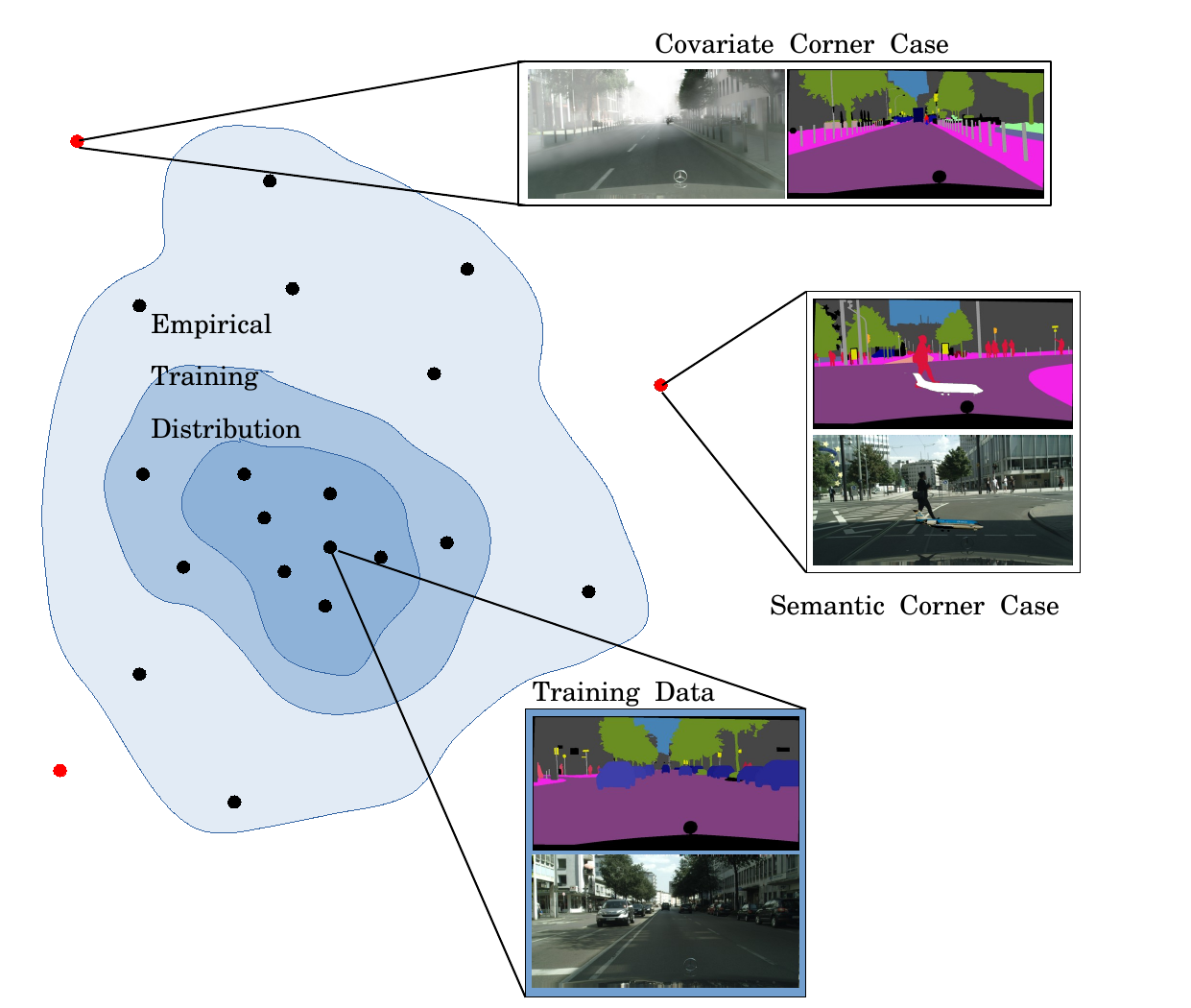}
    \vskip -0.3cm
    \caption{Illustration of Corner Cases in relation to the empirical training distribution. Training samples are marked with black circles, and Corner Cases are marked with red circles.
    We distinguish between the \localcc{} Corner Case and the \globalcc{} Corner Case, which are low-density regions in the training distribution, but have different data properties. 
    }
    \vskip -0.4cm
    \label{fig:CornerCaseDefinition}
\end{figure}
This definition, with its dedicated levels, covers most eventualities that can occur in open-world driving scenarios from a sensor perspective to the constellation of objects and participants. 
However, defining scenarios required exact assumptions on the open-world situation, which cannot be assumed, and the definition does reflect the actual dataset coverage.
Given that almost every perception system employs machine learning models, the importance of what is represented by the dataset is a fundamental factor in the performance of such systems in corner-case-like situations. For example, \cite{Bogdoll2021} defines a traffic jam as an anomaly, while existing Level 3 systems (BMW and Daimler) are exclusively operating in these situations.

To address this gap and the contradictions evident in these example-based definitions,
we propose a novel perspective on these \CornerCases{} based on the data coverage.
In our work, we introduce two types of \CornerCases{} definitions that are defined via the marginal data distribution: \emph{\localcc{} \CornerCases{}}, which typically affect specific local areas of a sample and include novel object instances and semantically unknown classes, and \emph{\globalcc{} \CornerCases{}}, which encompass factors such as weather-related changes. 
By building on machine learning concepts that reflect the distribution gap, we propose a framework to detect these cases and outline an approach for further temporal processing.
In our experiments, we show the effectiveness of our framework. For a broader evaluation, we introduce a novel variant of the Lost and Found dataset.
Our contribution can be summarized as:
\begin{itemize}
    \item We derive a novel perspective on automotive \CornerCases, leading to a data-based definition.
    \item We introduce a framework for detecting perception-based \CornerCases, based on training data distributions.
    \item In our evaluation, we showcase the effectiveness of our framework, including its performance on newly introduced dataset variants (Foggy Lost and Found), which we provide together with an evaluation benchmark.
\end{itemize}

\section{Related Work}
In the context of dealing with unknown and rare situations, besides the field of \emph{automotive \CornerCases{}}, we review the field of \emph{open-world perception} and \emph{OOD detection}.

\textit{1) Automated Driving \CornerCases{}.}
In autonomous driving, rare events are known as \CornerCases{}. However, the definition varies across the literature and is often based on specific examples.
Early definitions \cite{TowardsCC} state that a \CornerCase{} exists "if there is a non-predictable relevant object/class in a relevant location." This ambiguous definition highlights the challenge of identifying \CornerCases{}, as it depends on both the present objects and their locations. 
To offer a clearer definition, \cite{Breitenstein2020} proposed a systematic approach that examines various levels of detail in driving scenarios. This definition connects the theoretical concept of a \CornerCase{} with practical applications, and \cite{Heidecker2021} utilizes it to integrate \CornerCases{} into machine learning model training. \cite{cc_guidance_on_detection} has expanded this approach to include additional data sources like LiDAR and Radar.
Building on definitions from \cite{Breitenstein2020}, \cite{Bogdoll2021} discusses various scenario description languages and their integration of \CornerCases{}. However, they primarily offer a high-level overview and provide examples for the categorization, which are inspired by scenarios that humans find difficult. These scenarios comprise different levels of description, including \textbf{pixel level}, \textbf{domain level}, \textbf{object level}, and \textbf{scene level}. An overview is given in \cref{tab:cornerCaseDescription} on the left, next to our definition, derived in \cref{sec:dataDrivenCCDefinition}.
Based on these categories, other works \cite{cc_metric_safty_aware, cc_metric_risk_rank_recall, cc_metric_safty_concerns, Breitenstein2023} quantify \CornerCases{} by assessing their impact on model performance metrics. 
These studies define \CornerCases{} by specific scenarios, such as pedestrians on the road, but do not sufficiently explore how to generalize models to unseen real-world examples.
\cite{Heidecker2024} provides a theoretical definition of machine learning based \CornerCases{}, using an abstract relevance weighting to derive a "Corner Case Score." Yet, the ambiguity of relevance weights and the dependence on specific models complicate the identification.

Overall, existing definitions and methods for \CornerCases{} are primarily example-driven, making classification subjective, or overly reliant on test metrics rather than being data-driven.

\begin{table*}[h!]
\centering
\caption{Comparison of example-based and our data-based \CornerCases{} definitions. Some \CornerCases{} (*) require (additional) behavior adaptation and temporal consideration due to limited visibility, since risk can be learned from data.}
\vskip -0.2cm
\renewcommand{\arraystretch}{1.2}
\setlength{\tabcolsep}{3pt}
\begin{tabular}{|>{\raggedright\arraybackslash}p{3.4cm}|>{\raggedright\arraybackslash}p{7cm}|>{\centering\arraybackslash}p{1.65cm}|>{\centering\arraybackslash}p{1.78cm}|}
\hline
\rowcolor{gray!30}
\textbf{Example base Corner Case} & \textbf{Description} & \textbf{\Localcc{} \CornerCase{}} & \textbf{\Globalcc{} \CornerCase{}}\\
\hline

\multicolumn{2}{|l|}{\cellcolor{red!20}\textbf{Scenario Level} — Patterns observed over an image sequence; requires scene understanding} &  & \cmark  \\
Anomalous Scenario & Not observed during training; \textbf{high potential for collision}. & (\cmark )* & (\cmark )* \\
Novel Scenario & Not observed in training; no increased collision potential. & (\cmark ) & (\cmark ) \\
Risky Scenario & Observed in training; still contains potential for collision. & * & * \\

\multicolumn{2}{|l|}{\cellcolor{orange!20}\textbf{Scene Level} — Non-conformity with expected patterns in a single image} & & \cmark\\
Collective Anomaly & Multiple known objects, but in an unseen quantity. & (\cmark) & \cmark \\
Contextual Anomaly & A known object in an unusual location. & (\cmark) & \cmark\\

\multicolumn{2}{|l|}{\cellcolor{yellow!20}\textbf{Object Level} — Instances not seen before} & \cmark & \\
Single-Point Anomaly & An unknown (novel) object. & \cmark & \\

\multicolumn{2}{|l|}{\cellcolor{lime!20}\textbf{Domain Level} — World model fails to explain observations} & & \cmark\\
Domain Shift & Large, constant shift in appearance but not in semantics. & & \cmark\\

\multicolumn{2}{|l|}{\cellcolor{green!20}\textbf{Pixel Level} — (Perceived) errors in data} & & \cmark\\
Local Outlier & One/few pixels outside the expected range. & & \cmark\\
Global Outlier & Many pixels outside the expected range. & & \cmark  \\

\hline
\end{tabular}
\label{tab:cornerCaseDescription}
\vskip -0.5cm
\end{table*}

\textit{2) Out-of-Distribution Detection.}
While for autonomous driving \CornerCases, the current literature provides no clear definition, in machine learning, the task of OOD \cite{Drenkow2022} detection refers to the identification of whether a sample is part of the training distribution or not. This field has been particularly investigated for classification tasks. 
Detection approaches often require the output of the classification head for the sample assignment. 
Often, neurons and weights of the trained network are employed and include techniques such as filtering for important neurons \cite{ahn_line_2023}, or weights \cite{sun_dice_2021}, clipping neuron values to reduce OOD-induced noise \cite{sun_react_2021}, or activation scaling \cite{xu2024scaling}. Other approaches based on uncertainty \cite{liang_enhancing_2020}, energy scores \cite{liu_energy-based_2020}, or latent space distances \cite{Schmidt2025c} also require parts of the classification head.

Only a few approaches do not require classification head properties, such as density estimation approaches \cite{ramalho2019densityestimationrepresentationspace}.
While detecting the membership of a distribution might be interesting for \CornerCases, its applicability is limited, as most methods are tied to the classification task.

\textit{3) Open-World Perception.}
Open-world perception tasks, such as open-set segmentation or anomaly segmentation, aim to handle unknown situations, particularly those involving novel objects.
Literature distinguishes between \emph{anomaly segmentation} (binary separation of known vs. unknown pixels), \emph{anomaly instance segmentation} (differentiation of individual unknown objects), and \emph{open-set panoptic segmentation} (joint segmentation of known and unknown instances). Independent of this taxonomy, methods can be \emph{assumption-free}, requiring no auxiliary OOD data, or \emph{assumption-based}, relying on \emph{additional supervision or model constraints}.

For \emph{anomaly segmentation}, several works \cite{Deli2024,Nayal2022,Grcic2023,Rai2023} extend Mask2Former \cite{Cheng2022} by a pixel-wise uncertainty function and thresholding to distinguish between known and unknown pixels. However, they mostly rely on OOD training to improve their uncertainty estimation. 
To identify further instances \cite{Gasperini2023} extended uncertainty thresholding by clustering to differentiate between different anomaly instances.

In \emph{open-world semantic segmentation}, ContMAV \cite{Sodano2024} leverages \emph{void} objects to train a contrastive embedding decoder. By fitting class-wise Gaussians and combining them with distances in the contrastive embedding, ContMAV identifies semantically unknown classes besides known classes.
To further provide instance information of known and unknown classes as \emph{open-world panoptic segmentation},
EOPSN~\cite{Hwang2021} and DDOSP~\cite{Xu2022} rely on pseudo-labeling unknown regions as \emph{void} during training, thereby learning to identify them during inference. 
UgainS \cite{Nekrasov2023} tackles open-world panoptic segmentation by building on RbA \cite{Nayal2022} and using uncertainty-based prompting of the Segment Anything model (SAM) \cite{kirillov2023segment}.
U3HS \cite{Gasperini2023} builds on Panoptic Deeplab~\cite{Cheng2020} and is, hence, a fully convolutional network. It uses a Dirichlet Prior Network to enhance uncertainty estimation and combines it with DBScan clustering for anomaly instance segmentation. Prior2Former (P2F)~\cite{Schmidt2025b} builds upon Prior Network and introduces a framework for mask vision transformers.

While several approaches exist in machine learning to detect unknown objects and determine if a sample is part of the training distribution, the relationship to practical scenarios remains unexplored. Automotive \CornerCases{} aim to express difficult scenarios, but do not consider the underlying training data of a perception model. \emph{In our work, we address this gap and provide a Corner Case framework depending on machine learning tasks reflecting data distributions.
}

\section{Data Driven Corner Case Definition} %
\label{sec:dataDrivenCCDefinition}
To establish a data-driven definition of \CornerCases, we first introduce the terms "target", describing all possible autonomous driving scenarios, and "observed", reflecting the collected training data.
Let $P_{T}(x_T)$ be the target distribution over $x_T \in \Omega$, where $x_T$ are real world data points.
However, the point $x_T$ is only observed indirectly over a sensor $S:\Omega \rightarrow \mathbb{R}^{d \times f}$, such that $x_O = S(x_T)$. Here, $d$ represents the dimension of a linearized observed sample $x_O$ and $f$ the feature dimension, i.e., if $x_O$ is an image, $d$ would be height$\times$width and $f = 3$.
Additionally, the observed sample $x_O$ has label $y \in \mathbb{L}$, where $\mathbb{L}$ represents the label space. Note that the label and the label space depend on the given sensor $S$.
Accumulating all observed points $x_O$ with their labels $y$ into the set $ (x_T,y) \in D_O$, leads to the definition of the observed distribution $P_{D_O}(x_O,y)$ over the set $D_O$.

In the autonomous driving context, $x_T$ could be a scene from a street, $x_O$ an image from that scene using the camera as a sensor, $y$ a semantic segmentation label of that corresponding image, and $D_O$ would be the set of all the images $x_O$ together with their labels $y$. For model training, $D_O$ is split into training, validation, and test sets.

Under the described setting, a \CornerCase{} is then a newly observed sample $\tilde{x}_O = S(\tilde{x}_T)$ with label $\tilde{y}$ such that $P_{D_O}(\tilde{x}_O,\tilde{y}) \approx 0$. The following definition summarizes the derivation:
\vspace{0.2em}
\begin{definition}{\textbf{Corner Case for Autonomous Driving.}}
    \label{def:CornerCase}
    Let $S:\Omega \rightarrow \mathbb{R}^{d \times f}$ be a sensor function that maps a real-world data point $x_t\in \Omega$ to the observable space $x_O \in \mathbb{R}^{d \times f}$. Let $y \in \mathbb{L}$ be the corresponding label. Let $D_O$ be the set of labels and observed points seen during training with an empirical distribution $P_{D_O}$. 
    Then, a newly observed point $\tilde{x}_O = S(\tilde{x}_T)$ with label $\tilde{y}\in \mathbb{L}$ is a Corner Case if and only if:
    \begin{equation}
        P_{D_O}(\tilde{x}_O,\tilde{y}) \approx 0
    \end{equation}
     
\end{definition}

Following the above definition, we identify two main \CornerCases{}: \emph{a) \localcc{} Corner Case} , \emph{b) \globalcc{} Corner Case}.

\textbf{\Localcc{} Corner Cases} occur when new objects are observed through the sensor, while the global input appearance of the observed sample remains consistent with the training distribution. The \localcc{} \CornerCase{}, usually arises \textbf{locally} affecting a sub-space $\overline{d}<d$ of a given sample $\tilde{x}$. Here, the new objects can be unknown derivatives of known categories, or of unknown categories. This implies that the marginal distribution of the label is zero at that new point, while the marginal over the whole training sample remains positive. We formalize this as follows.
\vspace{0.2em}
\begin{definition}{\textbf{\Localcc{} Corner Case}.}
Let the setting be as in Definition~\ref{def:CornerCase}.
A newly observed point $\tilde{x}_O = S(\tilde{x}_T)$ with label $\tilde{y}$ is a \emph{\localcc{} Corner Case} if and only if:
\begin{equation}
P_{D_O,Y}(\tilde{y}) \approx 0
, \quad
P_{D_O,X}(\tilde{x}_{\overline{d},O}) \approx 0, \quad
P_{D_O,X}(\tilde{x}_{d,O}) > 0 ,
\end{equation}
where $P_{D_O,Y}$ is the marginal distribution over labels and $P_{D_O,X}$ the marginal distribution over inputs.
\end{definition}
\vspace{0.2em}

\textbf{\Globalcc{} Corner Cases} affect the entire sensor data, which can be interpreted as a \textbf{global}-level distribution shift of the observed sample, while the semantic content itself remains covered by the training label space. This translates to the mathematical formulation that the marginal distribution over the observed samples is approximately zero. For example, a shift in weather conditions or lighting conditions could be a \globalcc{} \CornerCase.
\vspace{0.2em}
\begin{definition}{\textbf{\Globalcc{} Corner Case}.}
Let the setting be as in Definition~\ref{def:CornerCase}.
A newly observed point $\tilde{x}_O = S(\tilde{x}_T)$ with label $\tilde{y}$ is a \emph{\globalcc{} Corner Case} if and only if:
\begin{equation}
P_{D_O,X}(\tilde{x}_O) \approx 0
\quad \text{and} \quad
P_{D_O,Y}(\tilde{y}) > 0 ,
\end{equation}
where $P_{D_O,X}$ is the marginal distribution over inputs and $P_{D_O,Y}$ the marginal distribution over labels.
\end{definition}
\vspace{0.2em}
An important notion of this \CornerCase{} definition is that including any form of additional data in the observed set $D_O$ implies that this specific additional data is no longer a \CornerCase. This includes any form of OOD leakage. Further, \emph{training on different weather conditions would mean that these different weather conditions are no longer \CornerCases{}}. This implies that \CornerCases{} can be resolved, for example, by adding specific samples to the data set with active learning \cite{Schmidt2020,Schmidt2023} or creating expected perturbations via augmentations.

\begin{figure*}[h]
    \centering
    \includegraphics[width=0.96\linewidth]{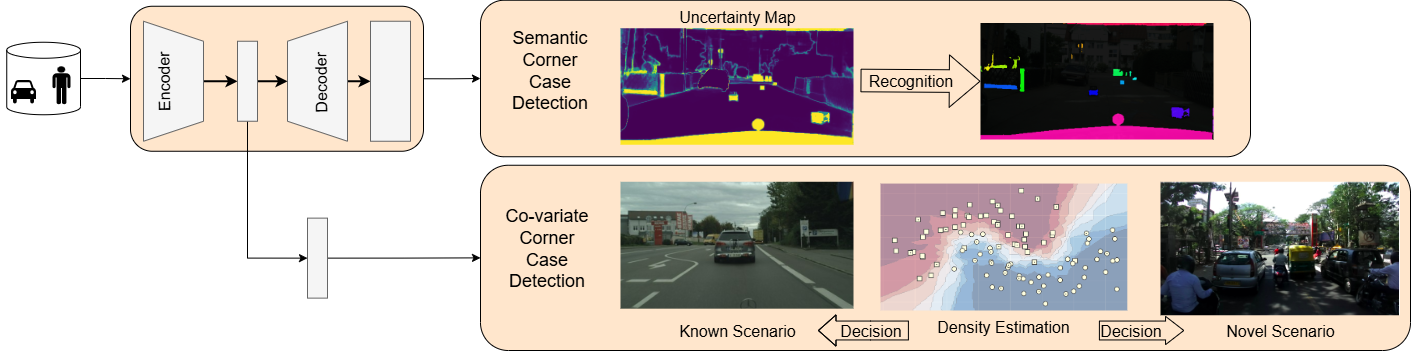}
    \vskip -0.3cm
    \caption{Data Driven Corner Case Framework: We aim to detect the \CornerCases{} in autonomous driving, which have been previously defined primarily on exemplar scenarios, based on membership of the training distribution. Therefore, we define a \localcc{} and a \globalcc{} Corner Case. For \localcc{} we employ open-world segmentation and extend it by a sample-level OOD detection to detect \globalcc{} \CornerCases{}.}
    \label{fig:Framework}
    \vskip -0.3cm
\end{figure*}

In \cref{fig:CornerCaseDefinition}, the \localcc{} and \globalcc{} \CornerCase{} is illustrated for a training on Cityscapes \cite{Cordts2016}, which does not include any adverse weather conditions or airplanes. In the illustration, we do not directly distinguish between the two \CornerCases{}; the distinction is implied by the reason for the data point to be in a low-density region of the empirical distribution over the training data.

Based on our two defined \CornerCases{}, all previously given example-based \CornerCases{} can be described. 
The \globalcc{} \CornerCase{} is capable of reflecting the \emph{Pixel} and \emph{Domain} level of the example-based definition, given its global and domain perspective. In addition, the \localcc{} \CornerCase{} addresses any kind of anomaly or novel class, which reflects the \emph{Object} and \emph{Scene} level \CornerCase{} of the example-based definition. Here it is important to note that Scene level \CornerCases{} include known objects, which either behave like unknown objects if a model is location dependent or are no corner case at all if a model is location invariant.
Yet, \emph{Scenario} level \CornerCases{} also include aspects that are not directly related to perception, such as general risk situations or cases of missing observability.
In these situations, the necessary context is not contained in the current data sample, either because it is unobservable or requires behavior adaptation. We refer to these as observability \CornerCases{}, where the missing information must be inferred from temporal context or prior knowledge due to behavior data attribution, e.g., “drive slowly near parked cars, since children can appear”. Since we focus on the sample-wise distribution definition, 
temporal considerations, which provide an explicit match, remain for future work.

\section{Corner Case Detection Framework}
\label{sec:Framework}
To detect the two introduced types of \emph{\CornerCases{}}, we propose a novel framework with separate detection branches for \localcc{} and \globalcc{} cases. 
Simplified, we assume that \globalcc{} \CornerCases{} are primarily global shifts affecting the entire sample, as they reflect scenarios, sensor perturbations, or weather conditions. This aligns well with the co-variate definition provided by the OpenOOD Framework \cite{Yang2022}.
In addition, novel semantic context is typically provided within a subspace of an image or sensor sample in the form of novel or unknown objects, and it does not occur uniformly throughout the sample. Therefore, we assume \localcc{} \CornerCases{} primarily in a subspace of a sample. 

Our proposed framework, as shown in \cref{fig:Framework}, targets \localcc{} \CornerCases{} with uncertainty-based open-world perception, specifically with an open-world segmentation approach, and \globalcc{} \CornerCases{} with an OOD detection module.
While OOD is a common practice for classification, its value for \CornerCases{} or complex tasks like segmentation remains unexplored. It should be noted that this framework provides flexibility in the concrete approaches and aims to provide a novel data-driven direction for \CornerCase{} recognition, rather than providing a concrete implementation. 

\textbf{\Localcc{} Corner Case.}
\Localcc{} \CornerCase{} occurs when unexpected or unknown objects appear in the scene, while the overall scene distribution remains familiar. Therefore, we build upon open-world panoptic segmentation approaches, incorporating an uncertainty mechanism without prior assumptions.
U3HS \cite{Gasperini2023} and P2F \cite{Schmidt2025b} have demonstrated a strong capability in this domain by identifying novel or anomalous objects through pixel-wise uncertainty estimation and clustering, making them ideal candidates.
Both approaches use evidential uncertainty based on a prior distribution, which is the Dirichlet distribution for the classification part of the semantic segmentation, defined by its contraction parameters $\boldsymbol{\kappa} = (\kappa_1, \kappa_2, \ldots, \kappa_C)$:
\begin{equation}
    \text{Dir}(\boldsymbol{p}|\boldsymbol{\kappa})= \frac{\Gamma(\sum_{k=1}^{K}\kappa_k)}{\prod_{k=1}^{K}\Gamma(\kappa_k)}\prod_{k=1}^{K}p_k^{\kappa_k-1}
\end{equation}
By using a (second-order) Dirichlet distribution over the (first-order) categorical distributions $\boldsymbol{p}$ a model can represent different types of uncertainty \cite{sale2023second}, which depend on the data distribution \cite{Gasperini2023,Schmidt2025b}, such that the uncertainty is defined by the magnitude of the concentration parameters $u = \frac{K}{\sum_{k=1}^K \kappa_k}$, reflecting the semantic label coverage.

Based on our \CornerCase{} definition, we apply these models with minor adaptation and focus our work on \globalcc{} and the combination of both \CornerCase.  

\textbf{\Globalcc{} Corner Case.}
For \globalcc{} \CornerCase, we take inspiration from OOD detection for classification, in which a score $s$ is calculated for every given input $X$ based on a scoring function $\delta$ to distinguish in-distribution (ID) from OOD data. For the decision, a threshold $\lambda$ is chosen to enable classification.
\begin{equation}
    \label{eq:classification_OOD}
    \text{Prediction}(X) = \begin{cases}
        \text{ID} & \delta(X) \geq \lambda \\
        \text{OOD} & \delta(X) < \lambda
    \end{cases}
\end{equation}
As open-world segmentation is well-suited for the \localcc{} \CornerCase{}, we aim to build on this task to provide a holistic \CornerCase{} detection. To enable a global shift detection, we are interested in a global score based on the pixel-wise prediction and uncertainties of a segmentation network. 
While such an ID and OOD distinction has been examined in OOD detection tasks for classification models, the application of this concept for a dense task like segmentation or detection remains non-trivial and has not been evaluated. As a result, there is no existing benchmark for evaluation or models for comparison.
We call this task ``Global OOD Detection'', emphasizing the global context of an image in a pixel-wise prediction task, and OOD Detection because the main goal is to distinguish between ID and OOD.
We propose two different directions, which do not require a specific (classification) network architecture.
The first uses latent space density estimation:
Therefore, we either employ a k-nearest neighbor method (KNN) to estimate density from distances to surrounding points \cite{ramalho2019densityestimationrepresentationspace} or Gaussian Mixture Models (GMM) to approximate the distribution density with Gaussians.
Second, we include an uncertainty statistic baseline aggregated over all pixels.

\textit{1) Density Estimation in the Latent Space.}
The density-based approaches are adapted from \cite{lee2018simpleunifiedframeworkdetecting} to work on encoder embeddings of our segmentation model. 
Most deep learning models have a classical encoder or backbone followed by a decoder for dense tasks (segmentation) or a specific head, for e.g., object detection. For an observed sample $X$ an encoder $f_{\text{Encoder}}$ projects the sample to latent space $Z=f_{\text{Encoder}}(X)$, which we can utilize for our density estimation approaches
\begin{equation}
Z = f_{\text{Encoder}}(X) \in \mathbb{R}^{C' \times H' \times W'}
\end{equation}
and average over the height $H'$ and width $W'$ in the latent space
\begin{equation}
Z' = g_{H',W'}(Z) := \frac{1}{H' W'} \sum_{i=1}^{H'} \sum_{j=1}^{W'} Z_{:,i,j} \in \mathbb{R}^{C'}
\end{equation}
resulting in a vector $Z'$ containing the average activation within a channel. 
On $Z'$ we perform a density estimation using either a nearest neighbor approach or a Gaussian-Mixture Model for density estimation.

For the estimation via KNN, we pass all the training images $x_O$ through the encoder $f_{\text{Encoder}}$ and store the mean feature activations in a set $\mathcal{Z}$. 
The OOD score for a new image $X$ with $Z' = g_{H',W'}(f_{\text{Encoder}}(X))$ being the mean activation of $Z=f_{\text{Encoder}}(X)$ is then
\begin{equation}
    s_{\text{KNN}} = -||Z'-Z_{(50)}||^2,
\end{equation}
where $Z_{(50)}$ is the $k$-th neighbor of $Z'$ in the set $\mathcal{Z}$.

We employ GMM by fitting it to the set $\mathcal{Z}$. Let $p_\mathcal{Z}$ be the density of the fitted model. The global OOD score is then
\begin{equation}
    s_{\text{GMM}} = p_\mathcal{Z}(Z'),
\end{equation}
where $Z'$ is again the mean feature activation of a given new image X.

\textit{2) Uncertainty Statistics.}
Segmentation models classically predict a pixel-wise uncertainty map $U$. Since we are interested in a single score per image, we employ the mean uncertainty over all pixels.
Given the uncertainty map $U \in \mathbb{R}^{H \times W}$ for an image, we compute the mean uncertainty:
\[
\bar{U} = \frac{1}{H \cdot W} \sum_{i=1}^H \sum_{j=1}^W U_{i,j}
\]
This aggregated score reflects the average model uncertainty over the entire image. A high mean uncertainty indicates that the model is less confident across large portions of the scene, which is typical for global distribution shifts such as different weather conditions or geographic domains.
For global OOD detection, we aggregate these pixel-wise uncertainties into a global uncertainty score.

\section{Evaluation}
To validate our framework, evaluate the individual detection branches for \localcc{} and \globalcc{} \CornerCase{} as well as the combination of both \CornerCase. We base the experiments regarding our framework on the prior works in the open-world perception domain U3HS \cite{Gasperini2023} and P2F \cite{Schmidt2025b}.
 Since these works provide extensive evaluation on detecting unknown instances, we provide only a short examination of \localcc{} \CornerCase{} and focus our evaluation on \globalcc{} \CornerCase{} and their combination.

\begin{figure*}[!ht]
    \centering
    \includegraphics[width=0.96\linewidth]{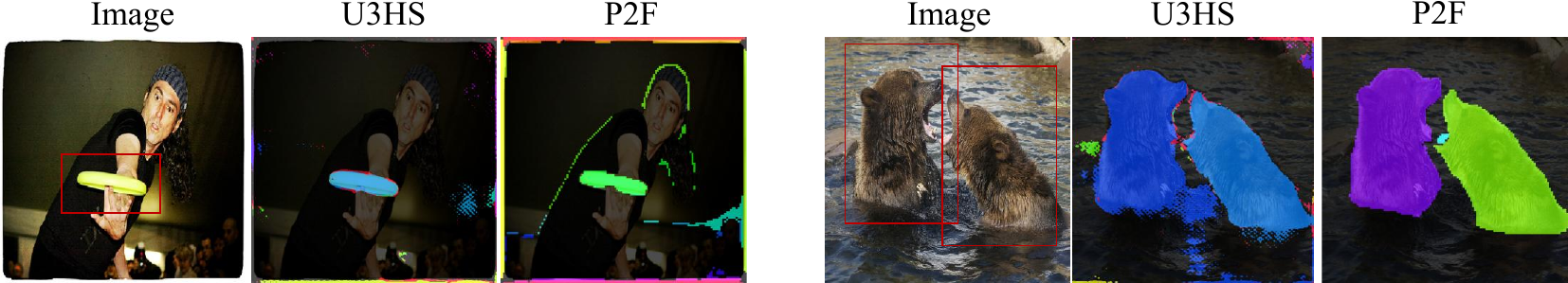}
    \vskip -0.3cm
    \caption{Evaluation of \localcc{} corner case: Detection of novel objects (frisbee and bear) of U3HS and P2F on COCO \cite{Lin2014}. It can be observed that both open-world segmentation approaches consistently detect instances of novel objects. Images are from \cite{Schmidt2025b}}
    \label{fig:localcc}
    \vskip -0.4cm
\end{figure*}

\textbf{\Localcc{} Corner Case.}
Within the proposed data-driven taxonomy (\cref{sec:dataDrivenCCDefinition}), a \localcc{} \CornerCase{} corresponds to situations in which novel or anomalous semantic content is present locally in the scene, while the global appearance distribution remains consistent with the training data. Formally, such cases arise when the marginal distribution over labels satisfies $P_{D_O,Y}(\tilde{y}) \approx 0$
for a given subset of the sample $(\tilde{x}_O, \tilde{y})$, while the marginal over inputs $P_{D_O,X}(\tilde{x}_{d,O})$ remains high. This distinguishes them from \globalcc{} \CornerCase, in which distributional shifts occur at the global image level.  

To operationalize \localcc{} \CornerCase{} detection within our framework, we build on top of the evidential uncertainty-based approaches \textbf{U3HS}~\cite{Gasperini2023} and \textbf{Prior2Former (P2F)}~\cite{Schmidt2025b} architectures. As mentioned in \cref{sec:Framework}, the prior assumption-free evidential uncertainty fits in our data-driven context. 
In \cref{fig:localcc} we can see that both approaches are able to detect the unknown classes frisebee and bear, which effectively addresses our \localcc{} \CornerCase{} problem.
Given the extensive evaluation on handling novel objects in the original works \cite{Gasperini2023,Schmidt2025b}, we limit our evaluation to a qualitative assessment of \localcc{} \CornerCase{} detection and focus further on \globalcc{} \CornerCases, while referring to \cite{Gasperini2023,Schmidt2025b} for further effectiveness proofs.

\textbf{\Globalcc{} Corner Case.}
The detection of \globalcc{} \CornerCases{} is not as easily reflectable by existing tasks as \localcc{} \CornerCases{}.
To evaluate the detection of \globalcc{} \CornerCase, we set up a detection benchmark and used Cityscapes (CS) \cite{Cordts2016} as the in-distribution (InD) dataset and trained the previously used U3HS \cite{Gasperini2023} and P2F as the basis for our framework implementation.
As OOD sources reflecting co-variate shifts and perturbations as \globalcc{} \CornerCases, we used Foggy \cite{SDV18} CS, Rainy CS \cite{Hu2019}, and ACDC \cite{sakaridis2024acdcadverseconditionsdataset} for adverse weather conditions like fog, rain, snow, and night. Furthermore, we utilized the Indian Driving Dataset (IDD) \cite{varma2018idddatasetexploringproblems} and A2D2 \cite{Geyer2020} to reflect major and minor domain shifts.
Lastly, we used dead pixels and noise to reflect sensor failure.

For a fair evaluation, we extend the OpenOOD benchmark for the task of \globalcc{} \CornerCase{} detection such that it includes the new datasets and models.
The benchmark provides \cite{Yang2022} a unified evaluation of the FPR (False Positive Rate), AUROC (Area Under the ROC Curve), AUPR In (Area Under the Precision–Recall curve for InD), and AUPR Out (Area Under the Precision–Recall curve for OOD) to measure a method's performance to recognize OOD data.  

\begin{table}[b]
\centering
\vskip -0.4cm
\caption{\Globalcc{} OOD Detection Results on multiple datasets using Cityscapes as In-Distribution.}
\vskip -0.2cm
\resizebox{\columnwidth}{!}{%
\begin{tabular}{l|l|cccc}
\toprule
Method & Dataset & FPR@95 $\downarrow$ & AUROC $\uparrow$ & AUPR\ IN $\uparrow$ & AUPR\ OUT $\uparrow$ \\
\midrule
GMM & Indian Driving & 0.79 & 99.78 & 98.13 & 99.97 \\
    & Foggy CS  & 64.11 & 84.50 & 85.68 & 81.95 \\
    & Rainy CS  & 4.53 & 99.07 & 98.83 & 99.29 \\
    & A2D2 & 0.59 & 99.86 & 99.24 & 99.97 \\
    & ACDC  & 0.33 & 99.89 & 99.51 & 99.97 \\
\midrule
KNN & Indian Driving & 0.85 & 99.76 & 97.82 & 99.97 \\
    & Foggy CS & 62.01 & 85.12 & 86.09 & 83.37 \\
    & Rainy CS & 5.38 & 98.92 & 98.55 & 99.19 \\
    & A2D2 & 0.85 & 99.85 & 98.97 & 99.97 \\
    & ACDC  & 0.20 & 99.90 & 99.45 & 99.97 \\
\midrule
U3HS & Indian Driving & 99.67 & 9.51 & 6.38 & 73.84 \\
     & Foggy CS & 95.08 & 50.22 & 48.84 & 51.13 \\
     & Rainy CS  & 97.11 & 44.58 & 38.67 & 52.60 \\
     & A2D2 & 97.51 & 37.02 & 12.14 & 78.60 \\
     & ACDC  & 97.90 & 29.52 & 14.14 & 69.49 \\
\midrule
P2F & Indian Driving & 93.96 & 70.16 & 27.10 & 92.44 \\
    & Foggy CS & 89.83 & 66.01 & 64.32 & 62.47 \\
    & Rainy CS & 97.90 & 49.65 & 47.37 & 53.62 \\
    & A2D2 & 100.00 & 18.50 & 10.13 & 68.83 \\
    & ACDC  & 99.93 & 44.87 & 28.80 & 72.07 \\
\bottomrule
\end{tabular}%
}
\vskip -0.2cm
\label{tab:globalood}
\end{table}
Using the framework extension, we evaluate in \cref{tab:globalood} the performance of our proposed GMM and KNN approaches using the U3HS decoder and the proposed uncertainty aggregation for both U3HS and P2F.
The results indicate that GMM and KNN effectively detect the IDD, Rainy CS, A2D2, and ACDC, demonstrating low FPR and high AUROC values. However, metrics for Rainy CS are slightly lower than those for the other datasets, likely due to differing camera sensors. Foggy CS exhibits notably poor performance across all metrics.
The uncertainty aggregation baselines underperform due to structural limitations in summarizing model confidence for global OOD detection. They compute global uncertainty by averaging pixel-wise uncertainty scores, which maintains the mean confidence but overlooks critical higher-order spatial statistics necessary to differentiate between co-variate shifts and InD variations. This averaging can mask significant distribution shifts, leading to an extremely low AUROC ($<50\%$) for U3HS and P2F on most datasets. From a probabilistic standpoint, both U3HS and P2F exhibit miscalibrated epistemic uncertainty due to a lack of explicit OOD training~\cite{guo2017calibration,ovadia2019trust}. Their uncertainty heads, designed for local anomalies, are sensitive to small, coherent unknown objects but struggle with global co-variate shifts, such as changes in weather or sensor domains. Additionally, the aggregation step assumes independence between pixel uncertainties, neglecting spatial correlations vital for effective global novelty detection.

\begin{figure*}[!ht]
    \centering
    \vskip -0.3cm
    \includegraphics[width=0.93\linewidth]{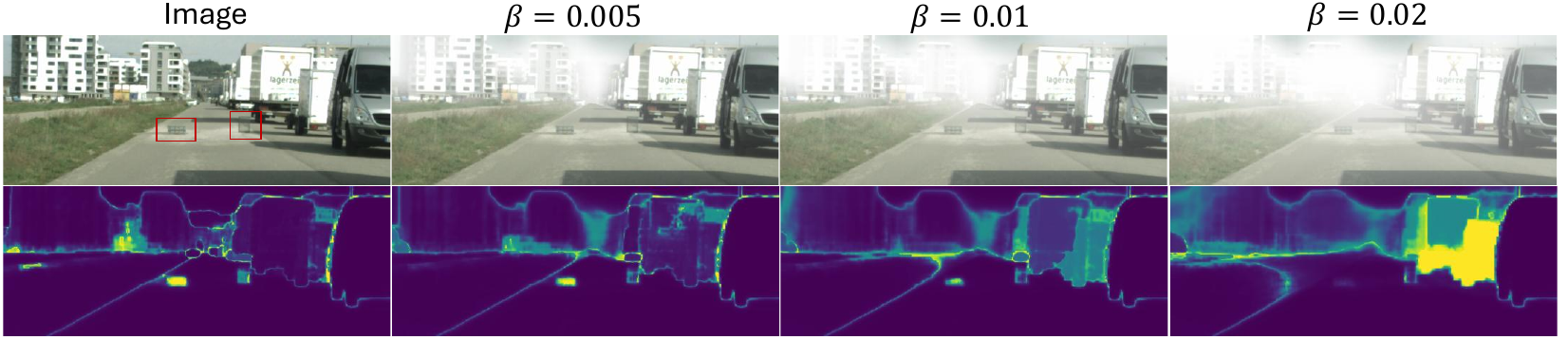}
    \vskip -0.35cm
    \caption{Visual Anomaly Segmentation performance of U3HS and P2F on L\&F and Foggy L\&F.}
    \vskip -0.2cm
    \label{fig:Task_Combination}
\end{figure*}

\begin{figure}
    \centering
    \vskip -0.2cm
    \includegraphics[width=\linewidth]{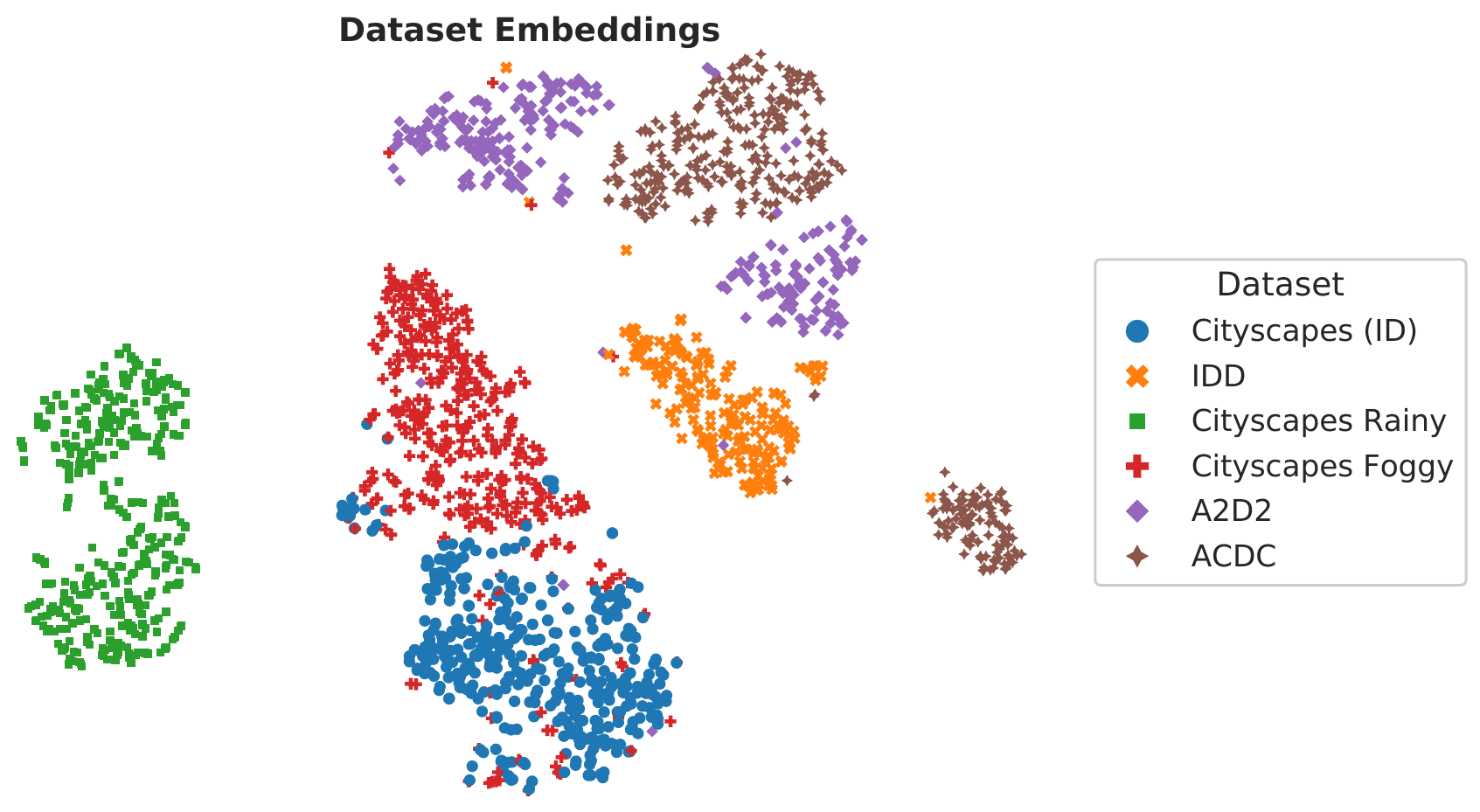}
    \vskip -0.4cm
    \caption{TSNE-Plot of OOD Datasets considering CS as InD.}
    \vskip -0.5cm
    \label{fig:tsne_datasets}
\end{figure}

Furthermore, we qualitatively examine the latent space of the datasets used in \cref{tab:globalood} in \cref{fig:tsne_datasets}. For this purpose, we randomly sample 100 images $X$ of each dataset and parse these images through our model, resulting in data points $Z'=g_{H',W'}(f_{\text{Encoder}}(X))$. We reduce dimensionality to 50 dimensions using Principal Component Analysis, and then use t-SNE Embeddings to further reduce the dimensions to two. 
The OOD datasets are clearly distinct from the ID data, which underscores the capabilities of our density estimation methods in capturing this separation. Only Foggy CS has an overlap with CS for a small number of data points, reflecting the lower scores in \cref{tab:globalood}. 

This overlap and low detection scores might be misleading, as the perturbation might not affect the model's performance and is therefore only a less critical \CornerCase. To evaluate this assumption, we examine three different fog intensities in \cref{tab:foglevels} and compare model performance (mIoU) with the \CornerCase{} detection abilities (FPR/AUROC) and their correlation. 
It can be seen that light fog (0.005 to 0.01) is difficult to detect but also has a minor impact on the mIoU of the models. In contrast, 0.02 has a significant impact on the model performance and is also much easier to detect. Yet the correlation of our detection indicates that the detection abilities increase with a higher performance impact.
This aligns with our \CornerCase{} definition that light fog is only slightly distinguishable from ID data, hence the empirical distribution at these images is not close to 0, resulting in a bad detection rate by the nature of the problem.

\begin{table*}[!ht]
\centering
\vskip -0.1cm
\caption{\Globalcc{} OOD Detection Results for Different Fogginess Levels in Cityscapes.}
\label{tab:foglevels}
\vskip -0.35cm
\begin{tabular}{@{} c|cc|cc c @{}}
\cmidrule[1.2pt](lr){1-5}\cmidrule[1.2pt](l){6-6}
\multicolumn{1}{c|}{\textbf{Fog Level}} &
\multicolumn{1}{c}{\textbf{FPR@95 $\downarrow$}} &
\multicolumn{1}{c|}{\textbf{AUROC $\uparrow$}} &
\multicolumn{1}{c}{\textbf{mIoU P2F $\uparrow$}} &
\multicolumn{1}{c}{\textbf{mIoU U3HS $\uparrow$}} &
\multicolumn{1}{c}{\textbf{Pearson corr. (GMM vs. mIoU)}} \\
\cmidrule[0.8pt](lr){1-5}\cmidrule[0.8pt](l){6-6} %

0.005& 84.25 & 67.81 & 75.79 & 58.81 &
\multirow{3}{*}{
  \begin{tabular}{@{}lcc@{}}
    & \textbf{P2F} & \textbf{U3HS} \\
    Cor FPR@95 & 97.02 & 98.89 \\
    Cor AUROC  & $-92.70$ & $-95.85$ \\
  \end{tabular}
} \\
0.01& 44.82 & 87.21 & 73.64 & 56.18 & \\
0.02& 6.96  & 98.50 & 68.49 & 51.89 & \\
\cmidrule(lr){1-5}\cmidrule(l){6-6} %
\cmidrule[1.2pt](lr){1-5}\cmidrule[1.2pt](l){6-6}
\end{tabular}
\vskip -0.45cm
\end{table*}
\begin{table}[b]
\centering
\vskip -0.25cm
\caption{Correlation metrics (Pearson and Spearman) between corruption severity and detection performance, for Gaussian noise and White pixels using GMM as detector.}
\vskip -0.25cm
\resizebox{\columnwidth}{!}{%
\begin{tabular}{llrrrr}
\toprule
Noise Type & Metric & \multicolumn{2}{c}{Pearson} & \multicolumn{2}{c}{Spearman} \\
\cmidrule(lr){3-4} \cmidrule(lr){5-6}
           &        & score       & p-value      & score       & p-value       \\
\midrule
Gaussian Noise & FPR@95 & -74.43 & $<10^{-9}$  & -97.34 & $<10^{-31}$ \\
               & AUROC  &  71.10 & $<10^{-8}$  &  99.11 & $<10^{-43}$ \\
\midrule
White Pixels   & FPR@95 & -91.23 & $<10^{-8}$  & -99.54 & $<10^{-20}$ \\
               & AUROC  &  90.70 & $<10^{-7}$  &  99.67 & $<10^{-21}$ \\
\bottomrule
\end{tabular}
}
\label{tab:correlations}
\vskip -0.1cm
\end{table}
Lastly, we evaluate detection of common sensor failures, \emph{white pixels} and \emph{Gaussian noise}, reflecting the sensor level \localcc{} \CornerCases{} with CS and respective noise augmentations. %

For the white-pixel corruption, we consider \textbf{20 box sizes} covering \([0.007, 0.119]\) of the image area; larger boxes saturate the detector (\(\mathrm{FPR@95} = 0.0\), \(\mathrm{AUROC} = 100.0\)), while smaller boxes are not reliably detected and are therefore excluded. 
For Gaussian noise, we vary the standard deviation over \textbf{50} equally spaced values in \([0.001, 0.01]\) with mean fixed at \(0\). 

For each corruption setting, we compute \(\mathrm{FPR@95}\) and \(\mathrm{AUROC}\) and report the \emph{Pearson} and \emph{Spearman} correlations (with p-values testing \(H_{0} : \rho = 0\)) between the corruption parameter (box area or \(\sigma\)) and each metric; see \cref{tab:correlations}. 
Both correlations indicate a strong monotonic dependence (in our results: \(\mathrm{FPR@95}\) decreases and \(\mathrm{AUROC}\) increases with increasing corruption severity), showing that our method not only detects such failures but also ranks their severity.

\textbf{Combination of Semantic and Co-variate Corner Cases}
From a distributional viewpoint, \localcc{} \CornerCases{} correspond to regions in the label space with negligible support, while \globalcc{} \CornerCases{} manifest as low-density regions in the input space. In real-world deployments, these two forms of distributional rarity often overlap, creating combined \CornerCases{} where both $P_{D_O,X}(\tilde{x}_O) \approx 0$ and $P{D_O,Y}(\tilde{y}) \approx 0$ hold simultaneously. 
For the autonomous driving application, we are interested in how the detection of anomalies changes in the presence of a \globalcc{} \CornerCase. To this end, we generate a new Lost \& Found Foggy dataset based on the Foggy CS generation \cite{SDV18} and the Lost \& Found Dataset (L\&F)~\cite{Pinggera2016}. The labels for the images remain unchanged. We will provide the dataset and instructions upon acceptance. 
We use the same dataset split as in \cite{Schmidt2025b} for evaluation and report the results of P2F and U3HS in \cref{tab:lf_obstacle_fog}.
We follow the classic (L\&F) evaluation scheme ~\cite{Pinggera2016}, report the AP and FPR95 for the local semantic \CornerCase{} (upper part) and the global co-variate \CornerCase{} detection in the lower part of \cref{tab:lf_obstacle_fog}. We see for GMM based on the U3HS backbone a correlation between fog intensity and \localcc{} and \globalcc{} \CornerCase{} detection for U3HS, while P2F is more robust.

\begin{table}[tbh]
\setlength{\tabcolsep}{2.4pt}
\centering
\vskip -0.3cm
\caption{L\&F Obstacle: anomaly segmentation (AP/FPR95) and GMM OOD detection (FPR95/AUROC) for different fog levels.}
\vskip -0.25cm
\resizebox{\columnwidth}{!}{%
\begin{tabular}{lcc|cc|cc|cc}

\toprule
& \multicolumn{2}{c}{No Fog} & \multicolumn{2}{c}{Fog $\beta=0.005$} & \multicolumn{2}{c}{Fog $\beta=0.01$} & \multicolumn{2}{c}{Fog $\beta=0.02$} \\
\cmidrule(lr){2-3}\cmidrule(lr){4-5}\cmidrule(lr){6-7}\cmidrule(lr){8-9}
Method & AP$\uparrow$ & FPR95$\downarrow$ & AP$\uparrow$ & FPR95$\downarrow$ & AP$\uparrow$ & FPR95$\downarrow$ & AP$\uparrow$ & FPR95$\downarrow$ \\
\midrule
U3HS & 55.8 & 26.1 & 51.7 & 45.4 & 49.6 & 43.8 & 50.1 & 40.8 \\
P2F  & 66.6 & 16.8 & 67.6 & 16.0 & 68.2 & 15.3 & 69.6 & 17.6 \\
\midrule
\cmidrule(lr){2-3}\cmidrule(lr){4-5}\cmidrule(lr){6-7}\cmidrule(lr){8-9}
\multicolumn{3}{l}{\textbf{\Globalcc{} \CornerCase{} Detection}} & FPR95$\downarrow$ & AUROC$\uparrow$ & FPR95$\downarrow$ & AUROC$\uparrow$ & FPR95$\downarrow$ & AUROC$\uparrow$ \\
\midrule
GMM & -- & -- & 92.3 & 60.5 & 84.6 & 70.3 & 47.8 & 88.2 \\
\bottomrule
\end{tabular}
}
\vskip -0.2cm
\label{tab:lf_obstacle_fog}
\end{table}

\section{Conclusion}
In our work, we presented a novel definition of autonomous driving corner cases (\CornerCases{}), namely \localcc{} and \globalcc{}, which aligns them to properties of the training data distribution and connects them to machine learning tasks. Based on our novel definition, we show that the classic example-based \CornerCases{} can be described and propose a framework for the detection. Our framework combined open-world segmentation to detect \localcc{} \CornerCases{} with a novel variant of out-of-distribution detection to identify \globalcc{} \CornerCases{}.
Especially, the detection of \globalcc{} \CornerCases{} requires a new task definition for which we provide a novel benchmark, which we make available. In addition, we provide a new dataset for the combination of \CornerCases{}.
In our experiments, we show the effectiveness of the different \CornerCases{} detection approaches as well as their combination. Especially for \globalcc{} \CornerCases, our proposed methods achieve an AUROC $>99\%$ in almost all cases. For \localcc{} \CornerCases{}, our framework integrates SotA uncertainty-based open-world networks for novel objects, enabling a holistic detection of both \globalcc{} and \localcc{} \CornerCases{}.

\bibliographystyle{IEEEtran}
\bibliography{main}

\end{document}